\documentclass{article}
\usepackage[preprint]{iclr2027_conference}
\usepackage{times}

\usepackage{amsmath,amsfonts,bm}

\def\figref#1{\hyperref[#1]{Fig.~\ref*{#1}}}
\def\Figref#1{\hyperref[#1]{Fig.~\ref*{#1}}}
\def\twofigref#1#2{\hyperref[#1]{Figs.~\ref*{#1}} and \hyperref[#2]{\ref*{#2}}}
\def\quadfigref#1#2#3#4{\hyperref[#1]{Figs.~\ref*{#1}}, \hyperref[#2]{\ref*{#2}}, \hyperref[#3]{\ref*{#3}} and \hyperref[#4]{\ref*{#4}}}
\def\secref#1{\hyperref[#1]{Sec.~\ref*{#1}}}
\def\Secref#1{\hyperref[#1]{Sec.~\ref*{#1}}}
\def\twosecrefs#1#2{\hyperref[#1]{Secs.~\ref*{#1}} and \hyperref[#2]{\ref*{#2}}}
\def\secrefs#1#2#3{\hyperref[#1]{Secs.~\ref*{#1}}, \hyperref[#2]{\ref*{#2}} and \hyperref[#3]{\ref*{#3}}}
\def\eqref#1{\hyperref[#1]{Eq.~(\ref*{#1})}}
\def\Eqref#1{\hyperref[#1]{Equation~(\ref*{#1})}}
\def\tabref#1{\hyperref[#1]{table~\ref*{#1}}}
\def\Tabref#1{\hyperref[#1]{Table~\ref*{#1}}}
\def\appref#1{\hyperref[#1]{appendix~\ref*{#1}}}
\def\Appref#1{\hyperref[#1]{Appendix~\ref*{#1}}}
\def\1{\bm{1}}

\DeclareMathAlphabet{\mathsfit}{\encodingdefault}{\sfdefault}{m}{sl}
\SetMathAlphabet{\mathsfit}{bold}{\encodingdefault}{\sfdefault}{bx}{n}

\usepackage{amsmath}
\usepackage{amssymb}
\usepackage{booktabs}
\usepackage{colortbl}
\usepackage{graphicx}
\usepackage{hyperref}
\usepackage{subcaption}
\usepackage{tikz}
\usepackage{url}
\usepackage{wrapfig}
\usetikzlibrary{arrows.meta,positioning}
\hypersetup{hidelinks}

\title{TailProp: Content-Adaptive Light- and Heavy-Tailed Propagation for Vision}

\author{Jiahao Kong\thanks{Corresponding author: Jiahao Kong (\texttt{jiahao\_kong@163.com})}\\
Shandong University\\
\And
Zihan Li\\
Shandong University}
\definecolor{TailPropHighlight}{RGB}{235,244,255}
\definecolor{PrelimBlue}{RGB}{36,96,180}

\definecolor{TailBlue}{RGB}{62,112,190}
\definecolor{TailGreen}{RGB}{74,150,116}
\definecolor{TailOrange}{RGB}{216,133,62}
\definecolor{TailInk}{RGB}{45,54,67}
\definecolor{TailSoft}{RGB}{244,248,252}

\tikzset{
  tailbox/.style={
    draw=TailBlue!70,
    fill=TailBlue!7,
    rounded corners=2pt,
    line width=0.45pt,
    align=center,
    inner sep=3pt,
    font=\sffamily\small
  },
  tailgreenbox/.style={
    draw=TailGreen!75,
    fill=TailGreen!8,
    rounded corners=2pt,
    line width=0.45pt,
    align=center,
    inner sep=3pt,
    font=\sffamily\small
  },
  tailorangebox/.style={
    draw=TailOrange!80,
    fill=TailOrange!9,
    rounded corners=2pt,
    line width=0.45pt,
    align=center,
    inner sep=3pt,
    font=\sffamily\small
  },
  tailarrow/.style={-{Latex[length=2.2mm]}, line width=0.55pt, draw=TailInk!80},
  tailthin/.style={line width=0.45pt, draw=TailInk!70}
}

\begin{document}
\raggedbottom
\maketitle

\begin{abstract}
Science-inspired vision models show that explicit propagation dynamics can provide structured and interpretable alternatives to conventional token mixing. Existing formulations, however, typically construct and adapt visual propagation within a particular dynamical family, while visual representations can require substantially different spatial interactions across samples, channels, and network stages. We explore \textbf{cross-regime adaptive propagation} and introduce \textbf{TailProp}, a hierarchical vision backbone built upon the \textbf{Tail Propagation Operator (TPO)}. TPO uses Gaussian and Cauchy stable-process propagators as complementary bases with rapidly decaying and heavy-tailed spatial influence, and predicts a content-conditioned channel-wise coefficient to adaptively combine them. Because this coefficient is spatially shared, the two responses are fused directly in the DCT domain with a single DCT/IDCT pair, yielding $O(N^{1.5})$ spatial mixing for square feature maps with $N=HW$ and fixed channel width. Across image classification, object detection, semantic segmentation, robustness, and cross-backbone restoration, TailProp consistently outperforms matched propagation baselines; TailProp-B reaches 84.4\% Top-1 accuracy on ImageNet-1K, 50.3/44.8 box/mask AP under the $3\times$ Mask R-CNN schedule, and 50.8\% mIoU on ADE20K. Controlled ablations further show that these gains are not explained by single-basis propagation, an additional same-family branch, or within-family adaptive order alone, supporting complementary two-basis propagation as an effective design principle for visual representation learning.
\end{abstract}

\section{Introduction}

Modern vision backbones differ in how they propagate information across spatial locations. Convolutional networks aggregate evidence through local filters and hierarchical receptive fields \citep{resnet,convnext}; vision transformers use self-attention to form pairwise token interactions \citep{vit,swin}; and recent state space models provide efficient long-range mixing through learned dynamical systems \citep{mamba,vmamba}. In parallel, science-inspired operators such as vHeat and WaveFormer show that explicit propagation dynamics can serve as efficient and interpretable inductive biases for visual modeling \citep{vheat,waveformer}.

A common pattern in science-inspired formulations is that the operator is organized around one dynamical family. Its parameters, diffusivity, damping, or even differential order may be learned or data-adaptive, so the issue is not whether prior work adapts. Instead, we ask whether visual representations also benefit from adapting across complementary propagation regimes. Since images, channels, and stages can require different ranges and decay profiles of spatial influence, a single propagation family may be an unnecessarily narrow modeling choice. Intuitively, compact structures whose useful correlations decay quickly may favor rapidly decaying propagation, whereas spatially distributed context or long-range co-occurrences may benefit from a slower-decaying influence profile, so assigning either regime a fixed semantic role would itself be unnecessarily restrictive.

\begin{figure}[t]
  \centering
  \includegraphics[width=\textwidth]{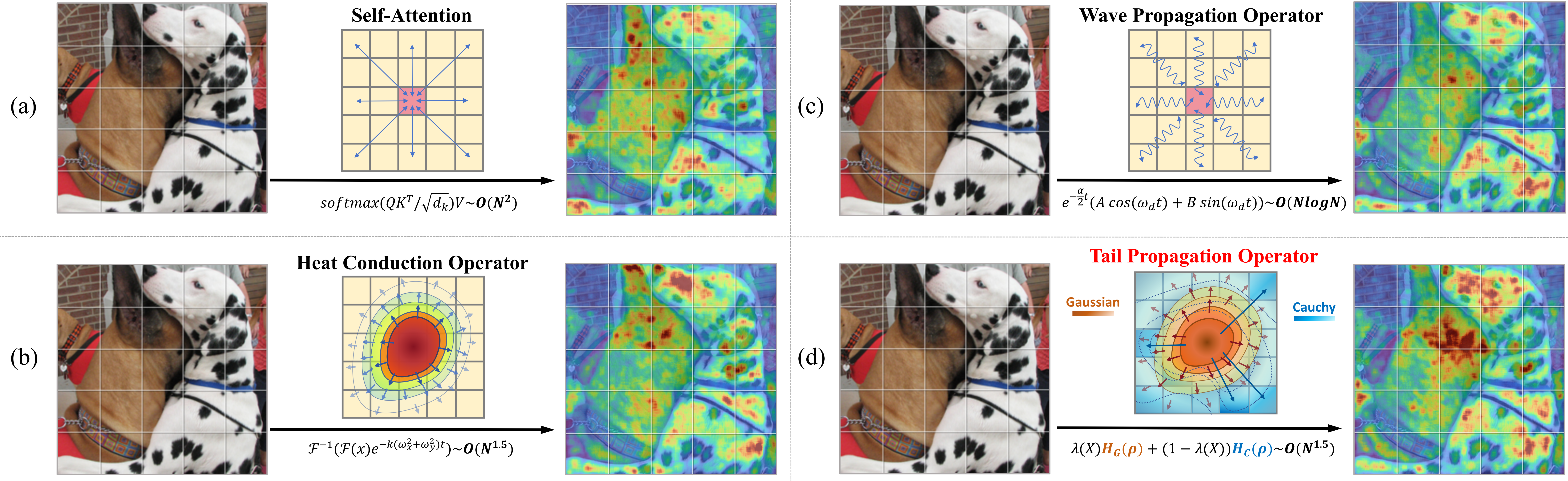}
  \caption{Comparison of visual propagation mechanisms. (a) Self-attention propagates information through pairwise token interactions with $O(N^2)$ complexity. (b) HCO performs heat-like diffusion, while (c) WPO introduces oscillatory wave propagation. (d) TPO adaptively combines light-tailed Gaussian and heavy-tailed Cauchy propagation through $\lambda(X)$, enabling complementary propagation behaviors with $O(N^{1.5})$ spatial scaling for fixed channel width. The right panels show the corresponding effective receptive fields.}
  \label{fig:overview}
\end{figure}

We study Gaussian and Cauchy propagation as two canonical symmetric stable-process cases \citep{samorodnitsky1994stable,kwasnicki2017ten}. Gaussian propagation is the $\alpha=2$ Brownian diffusion case, while Cauchy propagation is the $\alpha=1$ symmetric stable L\'evy process. Their contrast is clearest in the spatial domain: Gaussian kernels have rapidly decaying light-tailed influence, whereas Cauchy kernels have polynomial heavy tails and stronger relative far-field influence. As shown in \Figref{fig:overview}, TailProp makes this propagation choice content-adaptive.

We introduce the Tail Propagation Operator (TPO), which combines Gaussian and Cauchy propagators with an input-conditioned channel-wise coefficient. Because the coefficient is spatially shared, TPO fuses the two transfer responses in the DCT domain and applies the result with a single DCT/IDCT pair. For square feature maps with $N=HW$ and fixed channel width, this retains $O(N^{1.5})$ spatial mixing. TailProp stacks TPO Blocks and TPO Layers into a four-stage hierarchical backbone and instantiates Tiny, Small, and Base scales without changing the propagation principle.

We evaluate TailProp across ImageNet-1K classification, COCO detection and instance segmentation, ADE20K semantic segmentation, out-of-distribution robustness, cross-backbone restoration, and matched ablations. TailProp consistently achieves the best accuracy among the compared spectral propagation backbones across Tiny, Small, and Base scales, reaching 82.9/84.1/84.4\% Top-1 accuracy and extending the gains to COCO and ADE20K. Robustness, restoration transfer, and controlled ablations further show that the improvement generalizes beyond classification and is not explained by additional branch capacity or a single adaptive stable order.

Our contributions are threefold:
\begin{itemize}\setlength{\itemsep}{0pt}\setlength{\parsep}{0pt}\setlength{\topsep}{2pt}
\item We explore \textbf{cross-regime adaptive propagation} for science-inspired vision models, motivated by heterogeneous spatial interaction across samples, channels, and stages.
\item We introduce the \textbf{Tail Propagation Operator (TPO)} and TailProp backbone, which combine Gaussian and Cauchy propagators through content-conditioned channel-wise mixing and a fused spectral implementation requiring a single DCT/IDCT pair.
\item We demonstrate consistent gains across image classification, dense prediction, robustness, and cross-backbone restoration, while matched ablations isolate the benefit of complementary two-basis propagation and input-conditioned routing.
\end{itemize}

\section{Related Work}

\subsection{Vision Foundational Models}

Convolutional neural networks (CNNs) established the dominant paradigm for visual representation learning by exploiting locality and translation-equivariant inductive biases. Classic and efficient designs improved optimization, feature reuse, and deployability, while modern ConvNets further revisited scale, receptive field, and adaptive spatial aggregation \citep{alexnet,resnet,densenet,mobilenet,convnext,replknet,internimage}. Despite their efficiency and strong local modeling capability, long-range interactions in CNNs are still typically accumulated through kernels, hierarchy, or specialized aggregation rather than expressed as a direct global propagation rule.

Vision Transformers (ViTs) introduced self-attention as an explicit mechanism for modeling global dependencies among image tokens \citep{vit}. Hierarchical and efficient variants made transformer backbones more practical for dense and high-resolution vision tasks through pyramidal representations, windowed attention, cross-shaped attention, dual attention, cross-covariance attention, and simplified hierarchical designs \citep{swin,pvt,cswin,davit,xcit,hivit}. These designs substantially broadened global interaction modeling, but the cost and structure of token mixing remain central design constraints as image resolution grows.

More recently, state space models (SSMs) have emerged as another route to efficient long-range modeling. Multidimensional SSMs first showed that images and videos can be represented as continuous multidimensional signals, while Mamba-style selective state spaces provided a hardware-aware sequence modeling primitive \citep{s4nd,mamba}. Visual adaptations such as Vim and VMamba transfer these dynamics to images through bidirectional or two-dimensional selective scanning, and recent variants further reduce scanning cost, hybridize Mamba with attention, or question when SSM token mixing is necessary for vision \citep{vim,vmamba,efficientvmamba,mambavision,mambaout}. Collectively, these advances highlight a continuing shift from local feature aggregation toward efficient global interaction mechanisms; however, their propagation behavior is primarily determined by learned architectural operators, motivating complementary research that derives visual interaction rules from explicit and interpretable propagation priors.

\subsection{Science-Inspired Vision Models}

Scientific principles provide structured inductive biases for neural representation learning, ranging from biologically inspired spiking networks and nonequilibrium diffusion processes to PDE-guided visual modeling \citep{snnreview,ddpm,qbheat}. vHeat derives global semantic propagation from the heat equation, HcNet builds network components from heat-conduction dynamics, and WaveFormer employs an underdamped wave equation to model oscillatory and frequency-aware propagation \citep{vheat,hcnet,waveformer}. Related studies have also integrated anisotropic and reaction-diffusion processes into learned visual systems, further demonstrating the potential of explicitly structured diffusion dynamics \citep{metzger2023anisotropic,rao2023reactiondiffusion}.

More recently, fractional and nonlocal formulations have further expanded this design space by relaxing classical diffusion dynamics and allowing the governing operators or differential orders themselves to vary \citep{lfrd2,fractionalattention}. Learnable Fractional Reaction-Diffusion Dynamics predicts fractional differential orders for visual restoration, while Fractional Neural Attention models multiscale interactions through L\'evy diffusion governed by a fractional Laplacian \citep{lfrd2,fractionalattention}. Together, these studies show that propagation dynamics can be substantially enriched or adapted from data; nevertheless, existing science-inspired approaches predominantly instantiate or adapt one dynamical family at a time, leaving how to jointly exploit complementary propagation regimes for heterogeneous visual representations comparatively underexplored \citep{vheat,hcnet,waveformer,lfrd2,fractionalattention}. TailProp addresses this gap by enabling visual representations to adaptively draw on complementary propagation regimes rather than committing to a single dynamical family.

\section{Method}
\label{sec:method}
\subsection{Preliminaries: Gaussian and Cauchy Propagation}
\label{sec:preliminaries}

Let $u(\mathbf{x},t)$ denote a scalar field evolving over a two-dimensional domain, where $\mathbf{x}=(x,y)$ and $t$ denotes propagation time. A broad family of symmetric stable propagation processes can be described through the fractional diffusion equation \citep{metzler2000random,kwasnicki2017ten,samorodnitsky1994stable}
\begin{equation}
\frac{\partial u(\mathbf{x},t)}{\partial t}
=
-\kappa(-\Delta)^{\alpha/2}u(\mathbf{x},t),
\qquad 0<\alpha\leq2,
\label{eq:stable_pde}
\end{equation}
where $\kappa>0$ controls the propagation scale and $(-\Delta)^{\alpha/2}$ denotes the fractional Laplacian. Given the initial condition $u(\mathbf{x},0)=f(\mathbf{x})$, applying the Fourier transform $\mathcal{F}$ converts the equation into
\begin{equation}
\frac{\partial \widehat{u}(\boldsymbol{\omega},t)}
{\partial t}
=
-\kappa\|\boldsymbol{\omega}\|^{\alpha}
\widehat{u}(\boldsymbol{\omega},t),
\label{eq:stable_frequency_ode}
\end{equation}
where $\boldsymbol{\omega}=(\omega_x,\omega_y)$ denotes the spatial frequency. Solving the resulting ordinary differential equation gives
\begin{equation}
u(\mathbf{x},t)
=
\mathcal{F}^{-1}
\left[
\widehat{f}(\boldsymbol{\omega})
e^{-\kappa t\|\boldsymbol{\omega}\|^{\alpha}}
\right].
\label{eq:stable_solution}
\end{equation}

\begin{figure}[t!]
  \captionsetup{type=figure}
  \centering
  \includegraphics[width=\textwidth]{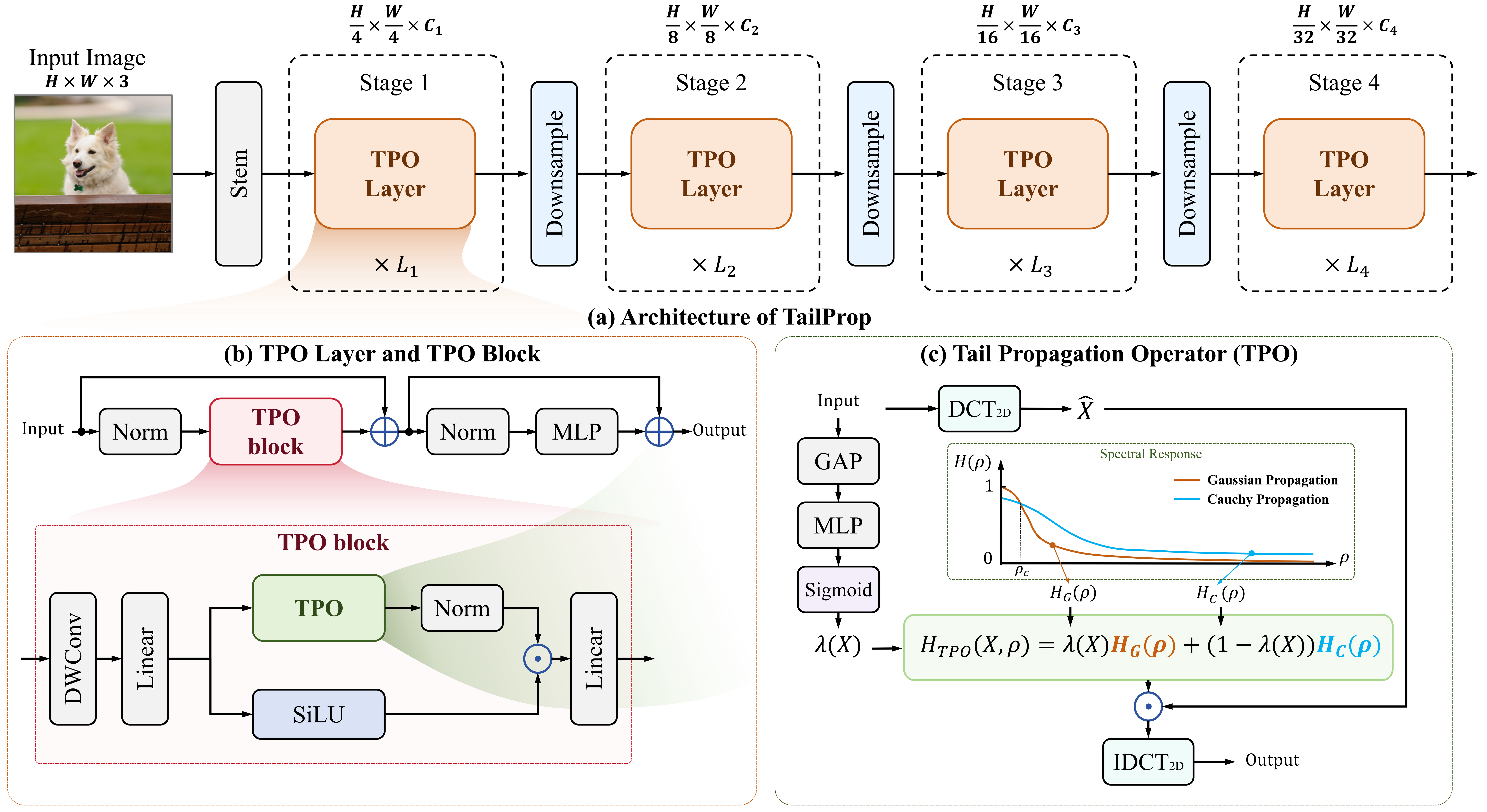}
  \caption{Overview of TailProp. (a) TailProp follows a four-stage hierarchical architecture built from TPO Layers. (b) Each TPO Layer contains a residual TPO Block and an MLP branch, with the TPO Block combining propagation and SiLU gating branches through element-wise modulation. (c) The Tail Propagation Operator (TPO) adaptively mixes Gaussian and Cauchy spectral responses and applies the fused response with a single DCT/IDCT pair.}
  \label{fig:tailprop_overview}
\end{figure}

\begin{samepage}
The solution in \eqref{eq:stable_solution} highlights two canonical and widely studied cases. For $\alpha=2$, the process reduces to \textbf{Gaussian propagation}, corresponding to Brownian diffusion; for $\alpha=1$, it becomes \textbf{Cauchy propagation}, namely the isotropic symmetric $1$-stable L\'evy process \citep{samorodnitsky1994stable,kwasnicki2017ten}. Their spectral transfer functions are respectively
\begin{equation}
H_G(\boldsymbol{\omega})
=
e^{-\kappa_Gt\|\boldsymbol{\omega}\|^2},
\qquad
H_C(\boldsymbol{\omega})
=
e^{-\kappa_Ct\|\boldsymbol{\omega}\|}.
\label{eq:canonical_transfer}
\end{equation}
\end{samepage}

Although both define globally supported propagation, their spatial influence decays in fundamentally different ways: the Gaussian kernel decays rapidly with distance, whereas the Cauchy kernel exhibits polynomial heavy tails and thus retains relatively stronger distant influence. These canonical yet complementary regimes motivate modeling visual propagation without committing to a single stable order; the learned preference between them is examined empirically in \Secref{sec:tailprop_analysis}.

\subsection{TailProp}
\label{sec:tailprop}

TailProp combines Gaussian and Cauchy propagation through content-adaptive mixing.

\paragraph{Tail Propagation Operator (TPO).}

Given an input feature map $X\in\mathbb{R}^{H\times W\times C}$, we extend the propagation processes in \Secref{sec:preliminaries} along the channel dimension. Since visual features are bounded rectangular signals, we adopt a Neumann boundary and use two-dimensional DCT/IDCT spectral propagation under the cosine-basis correspondence \citep{dct,vheat}. Further derivation and implementation details are provided in \Appref{app:dct_details}.

Let $\rho_{mn}=\omega_m^2+\omega_n^2$ denote the squared discrete spatial-frequency magnitude. The Gaussian and Cauchy propagation responses become
\begin{equation}
H_G(\rho_{mn})
=
e^{-\kappa_Gt\rho_{mn}},
\qquad
H_C(\rho_{mn})
=
e^{-\kappa_Ct\sqrt{\rho_{mn}}},
\label{eq:discrete_responses}
\end{equation}
where $\kappa_G$ and $\kappa_C$ are positive learnable propagation scales. TPO combines the two responses through a content-dependent coefficient $\lambda(X)$, yielding
\begin{equation}
H_{\mathrm{TPO}}(X,\rho)
=
\lambda(X)H_G(\rho)
+
[1-\lambda(X)]H_C(\rho)
\label{eq:tailprop_transfer}
\end{equation}
where $\lambda(X)\in(0,1)^C$ is a content-conditioned, channel-wise mixing coefficient predicted from the input feature; its exact parameterization is given in \eqref{eq:content_gate} in \Secref{sec:content_adaptive_tail_dynamics}.
The propagated feature is consequently obtained as
\begin{equation}
Y=
\operatorname{IDCT}_{2D}
\left(
H_{\mathrm{TPO}}(X,\rho)
\odot
\operatorname{DCT}_{2D}(X)
\right).
\label{eq:tailprop_operator}
\end{equation}

Because the channel-wise mixture coefficient is spatially shared, the Gaussian and Cauchy responses can be fused before the inverse transform, so \eqref{eq:tailprop_operator} needs one DCT/IDCT pair. For a feature map with $C$ channels, the separable matrix-DCT implementation has complexity $O(C(H^2W+HW^2))$; for square feature maps with $N=HW$ and fixed channel width, this gives $O(N^{1.5})$ spatial scaling. Details are in \Appref{app:matrix_dct}.

\paragraph{Content-Adaptive Tail Dynamics.}
\label{sec:content_adaptive_tail_dynamics}

Visual propagation requirements can vary across input samples, feature channels, and network stages. Global average pooling summarizes the current feature into a channel descriptor, from which a lightweight MLP predicts
\begin{equation}
\lambda(X)
=
\sigma\!\left(
\operatorname{MLP}
(\operatorname{GAP}(X))
\right),
\qquad
\lambda(X)\in(0,1)^C.
\label{eq:content_gate}
\end{equation}
The resulting coefficient depends on both the input sample and the feature channel while remaining spatially shared, avoiding location-wise routing. Its sample-dependent behavior is analyzed in \Secref{sec:tailprop_analysis} and \Figref{fig:gate_cases}.

\paragraph{TailProp Model.}

We construct a four-stage TailProp backbone with a convolutional stem and stages at $1/4$, $1/8$, $1/16$, and $1/32$ resolution. Each TPO Layer contains a residual TPO Block and an MLP branch. Within the TPO Block, depth-wise convolution and linear projection produce propagation and gating branches, which are processed by TPO and SiLU before element-wise fusion.

We instantiate three model scales, \textbf{TailProp-T, TailProp-S, and TailProp-B}, by varying stage depths and channel dimensions while keeping the propagation formulation unchanged. Detailed architectural configurations are provided in \Appref{app:tailprop_arch_configs}, and the key design choices are further discussed in \Secref{sec:tailprop_discussion}.

\subsection{Discussion}
\label{sec:tailprop_discussion}

\noindent We keep the discussion brief and focus on the key design choices underlying TailProp.

\textbf{Why are Gaussian and Cauchy propagation suitable for visual representation learning?}
Gaussian and Cauchy propagation provide complementary decay profiles: one induces a faster-decaying response, while the other preserves relatively stronger distant influence. Visual feature maps can require both behaviors across different channels and stages, so the two-basis formulation provides a broader response space than committing to a single decay profile.

\textbf{Why use two propagation bases instead of a single adaptive stable order?}
A single stable order imposes one order-dependent response over the frequency range, even when that order is conditioned on the input. In contrast, mixing Gaussian and Cauchy bases produces a frequency-dependent effective response that is not generally reducible to a single stable order; the matched Adaptive-$\alpha$ ablation in \Tabref{tab:ablation} empirically isolates this distinction.

\textbf{What advantages does TailProp offer over self-attention and single-family physics-inspired operators?}
Self-attention models pairwise affinities, which is powerful but quadratic in token count. TailProp avoids pairwise affinities and, unlike single-family physics-inspired operators, adapts across complementary stable-process responses rather than only within one dynamical family \citep{vheat,waveformer}. This gives global mixing with $O(N^{1.5})$ spatial scaling for square feature maps with $N=HW$ and fixed channel width.

\begin{center}
  \captionsetup{type=table}
  \caption{Main ImageNet-1K classification results. Best entries within each scale group are shown in \textbf{bold}. Lower is better for Params/FLOPs, and higher is better for throughput/accuracy. Throughput is measured on NVIDIA A100 80GB GPUs.}
  \label{tab:imagenet}
  \scriptsize
  \resizebox{\textwidth}{!}{%
  \begin{tabular}{lccccc}
    \toprule
    Method & Resolution & \#Params & FLOPs & Throughput (img/s) & Top-1 Acc. (\%) \\
    \midrule
    \multicolumn{6}{l}{\textit{Tiny / roughly 26--30M}} \\
    Swin-T~\citep{swin} & $224^2$ & 28M & 4.6G & 1242 & 81.3 \\
    ConvNeXt-T~\citep{convnext} & $224^2$ & 29M & 4.5G & 1198 & 82.1 \\
    Vim-S~\citep{vim} & $224^2$ & \textbf{26M} & 5.3G & 811 & 81.4 \\
    VMamba-T~\citep{vmamba} & $224^2$ & 30M & 4.9G & \textbf{1686} & 82.6 \\
    vHeat-T~\citep{vheat} & $224^2$ & 29M & 4.6G & 1514 & 82.2 \\
    WaveFormer-T~\citep{waveformer} & $224^2$ & 29M & \textbf{4.4G} & 1560 & 82.5 \\
    \rowcolor{TailPropHighlight}
    TailProp-T (Ours) & $224^2$ & 29M & 4.5G & 1468 & \textbf{82.9} \\
    \midrule
    \multicolumn{6}{l}{\textit{Small / roughly 50M}} \\
    Swin-S & $224^2$ & \textbf{50M} & 8.7G & 720 & 83.0 \\
    ConvNeXt-S & $224^2$ & \textbf{50M} & 8.7G & 687 & 83.1 \\
    VMamba-S & $224^2$ & \textbf{50M} & 8.7G & 877 & 83.6 \\
    vHeat-S & $224^2$ & \textbf{50M} & 8.5G & 945 & 83.6 \\
    WaveFormer-S & $224^2$ & \textbf{50M} & \textbf{7.8G} & \textbf{1020} & 83.9 \\
    \rowcolor{TailPropHighlight}
    TailProp-S (Ours) & $224^2$ & \textbf{50M} & 8.0G & 927 & \textbf{84.1} \\
    \midrule
    \multicolumn{6}{l}{\textit{Base / roughly 68--98M}} \\
    Swin-B & $224^2$ & 88M & 15.4G & 456 & 83.5 \\
    ConvNeXt-B & $224^2$ & 89M & 15.4G & 439 & 83.8 \\
    Vim-B & $224^2$ & 98M & 19.0G & 294 & 83.2 \\
    VMamba-B & $224^2$ & 89M & 15.4G & 528 & 83.9 \\
    vHeat-B & $224^2$ & \textbf{68M} & 11.2G & 661 & 84.0 \\
    WaveFormer-B & $224^2$ & \textbf{68M} & \textbf{10.8G} & \textbf{719} & 84.2 \\
    \rowcolor{TailPropHighlight}
    TailProp-B (Ours) & $224^2$ & \textbf{68M} & 11.0G & 639 & \textbf{84.4} \\
    \bottomrule
  \end{tabular}}
\end{center}

\section{Experiment}

\subsection{Experimental Settings}
We evaluate TailProp on image classification, object detection and instance segmentation, semantic segmentation, out-of-distribution robustness, and cross-backbone generalization. For image classification, we use ImageNet-1K \citep{imagenet} as the primary benchmark. For dense prediction, we transfer classification-pretrained backbones to Mask R-CNN \citep{maskrcnn} on MS COCO 2017 \citep{coco} and UPerNet \citep{upernet} on ADE20K \citep{ade20k}. We further evaluate ImageNet-1K-pretrained classifiers on ImageNet-Sketch \citep{imagenetsketch} and ImageNet-A \citep{imageneta} for robustness, and study cross-backbone generalization with SwinIR-style restoration \citep{swinir} on Set12 \citep{dncnn}, McMaster \citep{mcmaster}, and LIVE1 \citep{live}. Unless otherwise stated, experiments are configured and throughput/FPS are measured on NVIDIA A100 80GB GPUs. Detailed training recipes and implementation settings are provided in \Appref{app:experimental_details}.

\subsection{Experimental Results}
\noindent\textbf{Image Classification.} The ImageNet-1K results are summarized in \Tabref{tab:imagenet}. Under comparable model sizes and FLOPs, TailProp obtains the highest Top-1 accuracy at all three scales. TailProp-T reaches 82.9\% with 4.5G FLOPs and 1468 images/s, outperforming VMamba-T, WaveFormer-T, and vHeat-T by 0.3, 0.4, and 0.7 points. The modest throughput gap despite comparable FLOPs reflects that FLOP counts do not capture the transform and memory-access overhead of our current explicit matrix-DCT backend, which is not yet kernel-fused or hardware-specialized. The advantage persists at larger scales: TailProp-S/B achieve 84.1\%/84.4\%, improving over the strongest prior propagation baseline by 0.2 points in both groups. These results indicate that Gaussian--Cauchy adaptive propagation improves recognition accuracy while retaining comparable computational cost within spectral backbones.

\noindent\textbf{Object Detection and Instance Segmentation.} We transfer classification-pretrained backbones to Mask R-CNN on MS COCO 2017, and report 1x and 3x results in \Tabref{tab:coco}. TailProp consistently improves both box and mask AP across scales. Under the 1x schedule, TailProp-T obtains 46.2 APb and 41.8 APm, exceeding WaveFormer-T by 0.4/0.3 points and vHeat-T by 1.1/0.6 points with nearly identical FLOPs. The 3x schedule shows the same trend, where TailProp-T/S/B achieve 47.8/43.1, 49.4/44.2, and 50.3/44.8 APb/APm. Although TailProp is slightly below the fastest spectral baselines in FPS, it clearly improves detection quality at matched computational cost, suggesting that the complementary propagation bias transfers to localization and instance-level recognition.

\begin{center}
  \captionsetup{type=table}
  \caption{Mask R-CNN~\citep{maskrcnn} object detection and instance segmentation results on MS COCO 2017~\citep{coco}. Best entries within each scale group are shown in \textbf{bold}. FLOPs are calculated with input size $1280\times800$; FPS is measured on NVIDIA A100 80GB GPUs.}
  \label{tab:coco}
  \scriptsize
  \resizebox{\textwidth}{!}{%
  \begin{tabular}{lcccccccc}
    \toprule
    Method & \multicolumn{4}{c}{Mask R-CNN 1x} & \multicolumn{4}{c}{Mask R-CNN 3x} \\
    \cmidrule(lr){2-5}\cmidrule(lr){6-9}
    & APb & APm & FPS & FLOPs & APb & APm & FPS & FLOPs \\
    \midrule
    Swin-T~\citep{swin} & 42.7 & 39.3 & 26.3 & 267G & 46.0 & 41.6 & 26.3 & 267G \\
    ConvNeXt-T~\citep{convnext} & 44.2 & 40.1 & 29.3 & \textbf{262G} & 46.2 & 41.7 & 29.3 & \textbf{262G} \\
    vHeat-T~\citep{vheat} & 45.1 & 41.2 & \textbf{32.7} & 272G & 47.2 & 42.4 & 32.7 & 272G \\
    WaveFormer-T~\citep{waveformer} & 45.8 & 41.5 & 32.1 & 270G & 47.4 & 42.6 & \textbf{33.0} & 270G \\
    \rowcolor{TailPropHighlight}
    TailProp-T (Ours) & \textbf{46.2} & \textbf{41.8} & 31.1 & 271G & \textbf{47.8} & \textbf{43.1} & 30.9 & 271G \\
    \midrule
    vHeat-S & 46.8 & 42.3 & 25.9 & 348G & 48.8 & 43.7 & 25.9 & 348G \\
    WaveFormer-S & 47.0 & 42.5 & \textbf{26.2} & \textbf{345G} & 49.0 & 43.9 & \textbf{26.3} & \textbf{345G} \\
    \rowcolor{TailPropHighlight}
    TailProp-S (Ours) & \textbf{47.4} & \textbf{42.8} & 24.7 & 346G & \textbf{49.4} & \textbf{44.2} & 24.6 & 346G \\
    \midrule
    vHeat-B & 47.7 & 43.0 & 20.2 & 432G & 49.7 & 44.3 & 20.2 & 432G \\
    WaveFormer-B & 47.9 & 43.2 & \textbf{20.4} & \textbf{431G} & 49.9 & 44.5 & \textbf{20.4} & \textbf{431G} \\
    \rowcolor{TailPropHighlight}
    TailProp-B (Ours) & \textbf{48.2} & \textbf{43.4} & 19.3 & 432G & \textbf{50.3} & \textbf{44.8} & 19.3 & 432G \\
    \bottomrule
  \end{tabular}}
\end{center}

\noindent
\begin{minipage}[t]{0.50\textwidth}
  \vspace{0pt}
  \textbf{Semantic Segmentation.} We evaluate semantic segmentation with UPerNet on ADE20K, and summarize the results in \Tabref{tab:ade20k}. TailProp obtains the highest mIoU at all three scales. TailProp-T reaches 47.8 mIoU with 234G FLOPs, improving over WaveFormer-T and vHeat-T by 0.4 and 0.9 points. TailProp-S/B achieve 50.0/50.8 mIoU, exceeding WaveFormer-S/B with comparable FLOPs and supporting the transfer of TPO to dense scene parsing.

  The improvement remains consistent from Tiny to Base while keeping computational cost close to compared spectral backbones, suggesting that the representation gains of TPO extend from image-level recognition to dense pixel-level prediction rather than being confined to classification.
\end{minipage}
\hfill
\begin{minipage}[t]{0.48\textwidth}
  \vspace{0pt}
  \centering
  \captionsetup{type=table,font=footnotesize}
  \caption{UPerNet~\citep{upernet} results on ADE20K~\citep{ade20k}. Best entries within each scale group are shown in bold.}
  \label{tab:ade20k}
  \scriptsize
  \setlength{\tabcolsep}{2.4pt}
  \renewcommand{\arraystretch}{0.82}
  \resizebox{\textwidth}{!}{%
  \begin{tabular}{@{}lccc@{}}
    \toprule
    Method & mIoU & FPS & FLOPs \\
    \midrule
    Swin-T~\citep{swin} & 44.4 & 31.8 & 237G \\
    ConvNeXt-T~\citep{convnext} & 46.0 & \textbf{37.8} & 235G \\
    vHeat-T~\citep{vheat} & 46.9 & 36.7 & 235G \\
    WaveFormer-T~\citep{waveformer} & 47.4 & 36.9 & \textbf{233G} \\
    \rowcolor{TailPropHighlight}
    TailProp-T (Ours) & \textbf{47.8} & 36.4 & 234G \\
    \midrule
    vHeat-S & 49.1 & 26.1 & 254G \\
    WaveFormer-S & 49.8 & \textbf{26.4} & \textbf{252G} \\
    \rowcolor{TailPropHighlight}
    TailProp-S (Ours) & \textbf{50.0} & 25.6 & 253G \\
    \midrule
    vHeat-B & 49.6 & 23.6 & 293G \\
    WaveFormer-B & 50.5 & \textbf{23.8} & \textbf{290G} \\
    \rowcolor{TailPropHighlight}
    TailProp-B (Ours) & \textbf{50.8} & 23.2 & 291G \\
    \bottomrule
  \end{tabular}}
\end{minipage}

\noindent\textbf{Robustness Evaluation.} To assess robustness under distribution shift, we evaluate ImageNet-1K-pretrained classifiers on ImageNet-Sketch \citep{imagenetsketch} and ImageNet-A \citep{imageneta} without additional finetuning. As shown in the left part of \Tabref{tab:robustness_cross}, TailProp-B reaches 23.1 and 37.2 Top-1 accuracy, outperforming WaveFormer-B by 0.4/0.3 points and vHeat-B by 0.5/0.4 points. This consistent gain indicates that TPO improves robustness under sketch-style and naturally adversarial shifts.

\begin{center}
  \captionsetup{type=table}
  \caption{Robustness evaluation on ImageNet-Sketch~\citep{imagenetsketch} and ImageNet-A~\citep{imageneta}, and cross-backbone generalization on restoration benchmarks. Best entries are shown in \textbf{bold}.}
  \label{tab:robustness_cross}
  \scriptsize
  \setlength{\tabcolsep}{3.4pt}
  \resizebox{\textwidth}{!}{%
  \begin{tabular}{@{}lcc@{\hspace{0.7em}}lccc@{}}
    \toprule
    \multicolumn{3}{c}{Robustness Evaluation} &
    \multicolumn{4}{c}{Cross-Backbone Generalization} \\
    \cmidrule(lr){1-3}\cmidrule(lr){4-7}
    Model & ImageNet-Sketch & ImageNet-A & Model & Set12 & McMaster & LIVE1 \\
    & Top-1 & Top-1 & & PSNR ($\sigma=15$) & PSNR ($\sigma=15$) & PSNR ($q=40$) \\
    \midrule
    Swin-B~\citep{swin} & 22.1 & 36.0 & DnCNN~\citep{dncnn} & 32.86 & 33.45 & 33.96 \\
    ConvNeXt-B~\citep{convnext} & 22.4 & 36.5 & SwinIR~\citep{swinir} & 33.33 & 35.55 & 34.61 \\
    vHeat-B~\citep{vheat} & 22.6 & 36.8 & vHeatIR~\citep{vheat} & 33.37 & 35.60 & 34.64 \\
    WaveFormer-B~\citep{waveformer} & 22.7 & 36.9 & WaveFormerIR~\citep{waveformer} & 33.42 & 35.62 & 34.63 \\
    \rowcolor{TailPropHighlight}
    TailProp-B (Ours) & \textbf{23.1} & \textbf{37.2} &
    TailPropIR (Ours) & \textbf{33.51} & \textbf{35.69} & \textbf{34.72} \\
    \bottomrule
  \end{tabular}}
\end{center}

\noindent\textbf{Cross-Backbone Generalization.} We instantiate TailPropIR in a SwinIR-style restoration backbone. \Tabref{tab:robustness_cross} shows 33.51/35.69/34.72 PSNR on Set12, McMaster, and LIVE1, outperforming vHeatIR and WaveFormerIR and suggesting that the Gaussian--Cauchy bias transfers beyond classification.

\subsection{TailProp Analysis}
\label{sec:tailprop_analysis}
\noindent
\begin{minipage}[t]{0.58\textwidth}
  \vspace{0pt}
  \textbf{Core Ablation.} We compare matched Gaussian-only, Cauchy-only, Fixed G+C, Learnable G+C, Dual Gaussian, Adaptive-$\alpha$, and TailProp variants; Learnable G+C uses input-independent channel-wise gates in each TPO layer, whose final average corresponds to Gaussian/Cauchy weights of 0.53/0.47. Rather than exhaustively sweeping a scalar fixed $\lambda$, Learnable G+C provides a more expressive input-independent control by learning channel-wise mixing coefficients in every TPO layer. \Appref{app:ablation_details} gives definitions.

  These controls disentangle three factors: propagation-basis identity, dual-branch capacity, and conditioning strategy. Together, they test whether the gain can be explained by simply adding a branch, learning a global mixture, or adapting within a single stable family.
\end{minipage}
\hfill
\begin{minipage}[t]{0.38\textwidth}
  \vspace{0pt}
  \centering
  \captionsetup{type=table,font=footnotesize}
\caption{Core ablation.}
\label{tab:ablation}
\scriptsize
\setlength{\tabcolsep}{2.6pt}
\renewcommand{\arraystretch}{0.86}
\resizebox{\linewidth}{!}{%
  \begin{tabular}{@{}lcc@{}}
    \toprule
    Variant & Top-1 Acc. (\%) $\uparrow$ & $\Delta$ (pt) \\
    \midrule
    Gaussian-only & 81.9 & $-1.0$ \\
    Cauchy-only & 81.8 & $-1.1$ \\
    Fixed G+C & 82.3 & $-0.6$ \\
    Learnable G+C & 82.2 & $-0.7$ \\
    Dual Gaussian & 81.9 & $-1.0$ \\
    Adaptive $\alpha$ & 82.2 & $-0.7$ \\
    \rowcolor{TailPropHighlight}
    TailProp (Ours) & \textbf{82.9} & \textbf{0.0} \\
    \bottomrule
  \end{tabular}}

\end{minipage}

\noindent \Tabref{tab:ablation} gives TailProp the best result (82.9), ahead of Fixed G+C (82.3), Learnable G+C/Adaptive-$\alpha$ (82.2), Gaussian-only/Dual Gaussian (81.9), and Cauchy-only (81.8). The single-basis and Dual Gaussian controls show that the gain comes from heterogeneous Gaussian--Cauchy complementarity rather than an extra branch. The near-balanced but weaker Learnable G+C baseline indicates that input-conditioned routing supplies the remaining improvement; Adaptive-$\alpha$ further shows that one adaptive order is insufficient. Together, these controls attribute the gain to complementary basis diversity and input-conditioned routing rather than extra branch capacity or adaptation within a single stable order.

\noindent\textbf{Mechanism Visualization.}
We examine TPO from two complementary views. \Figref{fig:gate_cases} shows what the trained model chooses: the input-conditioned, channel-wise gate varies across samples and stages, and the shared spectral panel directly compares the induced TPO responses. Because all audited sample-average values remain below 0.5, we describe samples as having relatively low, intermediate, or high $\lambda$. This view is sample-centric: it reflects how the model allocates mass between the two propagation bases across stages, not a fixed backbone-level preference. Put differently, Figure 3 summarizes where the learned mixture lands in practice, while the accompanying spectra show which response shapes those choices induce.

\Figref{fig:controlled_tpo} then isolates what changing $\lambda$ does by fixing the learned Stage-3 propagation scales and applying the same centered impulse through the exact DCT TPO path. Increasing $\lambda$ shifts the response toward the Gaussian basis, whereas lower $\lambda$ gives relatively stronger Cauchy far-field response on the log radial profile. The spatial maps and radial decay therefore agree on the same qualitative trend from two perspectives: higher $\lambda$ concentrates response more tightly near the source, while lower $\lambda$ preserves stronger sufficiently-far influence on the shared scale. Additional stage-wise diagnostics are provided in \Appref{app:mechanism_diagnostics}. Thus, the learned adaptation is better interpreted as continuous sample-dependent reweighting within the two-basis response space rather than hard switching between the two endpoints. In this sense, Figure 4 complements Figure 3 by holding the input fixed and varying only $\lambda$, so the contrast isolates propagation geometry rather than routing variability.

\newpage

\begin{center}
  \centering
  \captionsetup{type=figure}
  \includegraphics[width=0.90\textwidth]{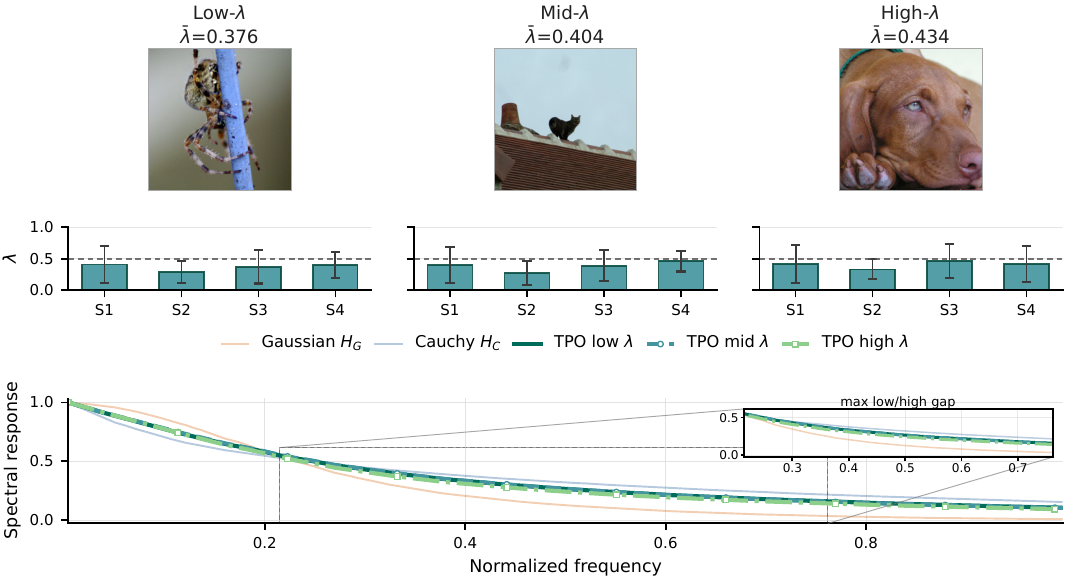}
  \caption{Sample-dependent TPO mixing preferences. The three examples are samples with relatively low, intermediate, and high $\lambda$ within the audited validation distribution. Bars show stage-wise $\lambda$, and the bottom panel compares their sample-conditioned TPO spectral responses. The inset enlarges the frequency interval automatically selected by maximal separation between the low- and high-$\lambda$ responses.}
  \label{fig:gate_cases}
\end{center}

\begin{center}
  \centering
  \captionsetup{type=figure}
  \includegraphics[width=0.92\textwidth]{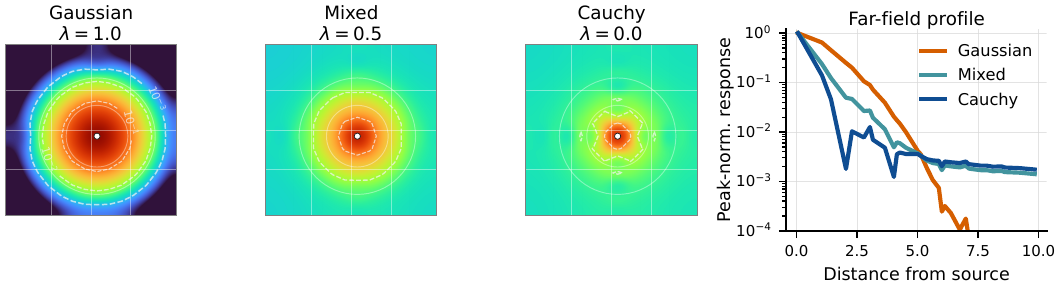}
  \caption{Controlled TPO propagation visualization. With learned Stage-3 scales and the centered impulse fixed, we vary only scalar $\lambda$ in the exact DCT-domain TPO. Spatial maps show peak-normalized response on a shared log scale with fixed contours, and the right panel reports radial decay.}
  \label{fig:controlled_tpo}
\end{center}

\section{Conclusion}
We presented TailProp, a science-inspired vision backbone for content-adaptive propagation across complementary dynamical regimes. Through the Tail Propagation Operator (TPO), TailProp mixes Gaussian and Cauchy stable-process bases with content-conditioned channel-wise gates, adapting between light- and heavy-tailed propagation while retaining a fused spectral path. Results across classification, dense prediction, robustness, cross-backbone transfer, and controlled ablations support this two-basis design and show that the gains are not explained by single-basis or single-order adaptation alone.

\section{Limitations and Future Work}
TailProp leaves several directions open. We study only Gaussian and Cauchy bases; broader stable families or learnable basis sets may capture additional regimes. Its gate is sample- and channel-conditioned but spatially shared to preserve one DCT/IDCT path, so efficient location-dependent routing remains open. The matrix-DCT implementation is not hardware-optimal, motivating fused transform kernels. Future work can extend cross-regime propagation to video, generation, multimodal learning, and embodied perception.

\section*{Reproducibility Statement}
We have made efforts to ensure that the results in this work are reproducible. The mathematical formulation and implementation of the Tail Propagation Operator are described in Sec.~3, with additional details on the DCT/IDCT formulation, boundary treatment, and fused matrix-DCT implementation provided in \Appref{app:dct_details} and \Appref{app:matrix_dct}. Detailed configurations of the TailProp-T/S/B architectures are reported in \Appref{app:tailprop_arch_configs}, while training recipes, optimization settings, data processing procedures, checkpoint selection, and downstream evaluation protocols are documented in \Appref{app:experimental_details}. \Appref{app:ablation_details} specifies the construction and purpose of all matched ablation variants, and additional implementation details are provided in \Appref{app:implementation_details}. Source code, configuration files, training and evaluation scripts, environment specifications, and instructions for reproducing the main experiments and analyses will be included in the supplementary material accompanying the submission.

\section*{AI Use Statement}
The core research idea and initial hypothesis of this work were conceived independently by the authors through literature study, experimental observations, and analysis of existing science-inspired vision models. Generative AI tools were not used to originate the central idea of cross-regime Gaussian--Cauchy propagation. After the initial research direction had been established, generative AI tools were used as assistive tools for literature organization, critical discussion and refinement of the research narrative, feedback on experimental design, code and script development, figure prototyping, and language editing of portions of the manuscript. AI-assisted suggestions were treated as preliminary inputs rather than authoritative results. All mathematical formulations, methodological decisions, experimental procedures, quantitative results, and scientific claims were independently checked and validated by the authors. AI-assisted code was inspected and tested before use, and cited literature and bibliographic information were manually verified against the original sources. The authors take full responsibility for the final content, results, and conclusions of this work.

\bibliographystyle{iclr2027_conference}
\bibliography{reference}

@inproceedings{resnet,
  title={Deep Residual Learning for Image Recognition},
  author={He, Kaiming and Zhang, Xiangyu and Ren, Shaoqing and Sun, Jian},
  booktitle={Proceedings of the IEEE Conference on Computer Vision and Pattern Recognition},
  pages={770--778},
  year={2016}
}

@inproceedings{convnext,
  title={A ConvNet for the 2020s},
  author={Liu, Zhuang and Mao, Hanzi and Wu, Chao-Yuan and Feichtenhofer, Christoph and Darrell, Trevor and Xie, Saining},
  booktitle={Proceedings of the IEEE/CVF Conference on Computer Vision and Pattern Recognition},
  pages={11976--11986},
  year={2022}
}

@inproceedings{replknet,
  title={Scaling Up Your Kernels to 31x31: Revisiting Large Kernel Design in CNNs},
  author={Ding, Xiaohan and Zhang, Xiangyu and Han, Jungong and Ding, Guiguang},
  booktitle={Proceedings of the IEEE/CVF Conference on Computer Vision and Pattern Recognition},
  pages={11963--11975},
  year={2022}
}

@inproceedings{vit,
  title={An Image Is Worth 16x16 Words: Transformers for Image Recognition at Scale},
  author={Dosovitskiy, Alexey and Beyer, Lucas and Kolesnikov, Alexander and Weissenborn, Dirk and Zhai, Xiaohua and Unterthiner, Thomas and Dehghani, Mostafa and Minderer, Matthias and Heigold, Georg and Gelly, Sylvain and Uszkoreit, Jakob and Houlsby, Neil},
  booktitle={International Conference on Learning Representations},
  year={2021}
}

@inproceedings{swin,
  title={Swin Transformer: Hierarchical Vision Transformer Using Shifted Windows},
  author={Liu, Ze and Lin, Yutong and Cao, Yue and Hu, Han and Wei, Yixuan and Zhang, Zheng and Lin, Stephen and Guo, Baining},
  booktitle={Proceedings of the IEEE/CVF International Conference on Computer Vision},
  pages={10012--10022},
  year={2021}
}

@inproceedings{pvt,
  title={Pyramid Vision Transformer: A Versatile Backbone for Dense Prediction Without Convolutions},
  author={Wang, Wenhai and Xie, Enze and Li, Xiang and Fan, Deng-Ping and Song, Kaitao and Liang, Ding and Lu, Tong and Luo, Ping and Shao, Ling},
  booktitle={Proceedings of the IEEE/CVF International Conference on Computer Vision},
  pages={568--578},
  year={2021}
}

@inproceedings{cswin,
  title={CSWin Transformer: A General Vision Transformer Backbone with Cross-Shaped Windows},
  author={Dong, Xiaoyi and Bao, Jianmin and Chen, Dongdong and Zhang, Weiming and Yu, Nenghai and Yuan, Lu and Chen, Dong and Guo, Baining},
  booktitle={Proceedings of the IEEE/CVF Conference on Computer Vision and Pattern Recognition},
  pages={12124--12134},
  year={2022}
}

@inproceedings{hivit,
  title={HiViT: A Simpler and More Efficient Design of Hierarchical Vision Transformer},
  author={Zhang, Xiaosong and Tian, Yunjie and Xie, Lingxi and Huang, Wei and Dai, Qi and Ye, Qixiang and Tian, Qi},
  booktitle={International Conference on Learning Representations},
  year={2023}
}

@inproceedings{mamba,
  title={Mamba: Linear-Time Sequence Modeling with Selective State Spaces},
  author={Gu, Albert and Dao, Tri},
  booktitle={Proceedings of the First Conference on Language Modeling},
  year={2024},
  url={https://openreview.net/forum?id=tEYskw1VY2}
}

@inproceedings{vim,
  title={Vision Mamba: Efficient Visual Representation Learning with Bidirectional State Space Model},
  author={Zhu, Lianghui and Liao, Bencheng and Zhang, Qian and Wang, Xinlong and Liu, Wenyu and Wang, Xinggang},
  booktitle={Proceedings of the Forty-first International Conference on Machine Learning},
  pages={62429--62442},
  year={2024},
  volume={235},
  series={Proceedings of Machine Learning Research},
  publisher={PMLR}
}

@inproceedings{vmamba,
  title={VMamba: Visual State Space Model},
  author={Liu, Yue and Tian, Yunjie and Zhao, Yuzhong and Yu, Hongtian and Xie, Lingxi and Wang, Yaowei and Ye, Qixiang and Jiao, Jianbin and Liu, Yunfan},
  booktitle={Advances in Neural Information Processing Systems},
  volume={37},
  year={2024},
  doi={10.52202/079017-3273}
}

@article{efficientvmamba,
  title={EfficientVMamba: Atrous Selective Scan for Light Weight Visual Mamba},
  author={Pei, Xiaohuan and Huang, Tao and Xu, Chang},
  journal={Proceedings of the AAAI Conference on Artificial Intelligence},
  volume={39},
  number={6},
  pages={6443--6451},
  year={2025},
  doi={10.1609/aaai.v39i6.32690}
}

@inproceedings{mambavision,
  title={MambaVision: A Hybrid Mamba-Transformer Vision Backbone},
  author={Hatamizadeh, Ali and Kautz, Jan},
  booktitle={Proceedings of the IEEE/CVF Conference on Computer Vision and Pattern Recognition},
  pages={25261--25270},
  year={2025}
}

@inproceedings{mambaout,
  title={MambaOut: Do We Really Need Mamba for Vision?},
  author={Yu, Weihao and Wang, Xinchao},
  booktitle={Proceedings of the IEEE/CVF Conference on Computer Vision and Pattern Recognition},
  pages={4484--4496},
  year={2025}
}

@article{snnreview,
  title={Deep Learning in Spiking Neural Networks},
  author={Tavanaei, Amirhossein and Ghodrati, Masoud and Kheradpisheh, Saeed Reza and Masquelier, Timothee and Maida, Anthony},
  journal={Neural Networks},
  volume={111},
  pages={47--63},
  year={2019},
  doi={10.1016/j.neunet.2018.12.002}
}

@inproceedings{ddpm,
  title={Denoising Diffusion Probabilistic Models},
  author={Ho, Jonathan and Jain, Ajay and Abbeel, Pieter},
  booktitle={Advances in Neural Information Processing Systems},
  volume={33},
  pages={6840--6851},
  year={2020}
}

@inproceedings{vheat,
  title={Building Vision Models upon Heat Conduction},
  author={Wang, Zhaozhi and Liu, Yue and Tian, Yunjie and Liu, Yunfan and Wang, Yaowei and Ye, Qixiang},
  booktitle={Proceedings of the IEEE/CVF Conference on Computer Vision and Pattern Recognition},
  pages={9707--9717},
  year={2025}
}

@inproceedings{hcnet,
  title={Efficient Visual Representation Learning with Heat Conduction Equation},
  author={Zhang, Zhemin and Gong, Xun},
  booktitle={Proceedings of the Thirty-Fourth International Joint Conference on Artificial Intelligence},
  pages={2431--2439},
  year={2025},
  doi={10.24963/ijcai.2025/271},
  url={https://www.ijcai.org/proceedings/2025/271}
}

@article{waveformer,
  title={WaveFormer: Frequency-Time Decoupled Vision Modeling with Wave Equation},
  author={Shu, Zishan and Wu, Juntong and Yan, Wei and Liu, Xudong and Zhang, Hongyu and Liu, Chang and Mao, Youdong and Chen, Jie},
  journal={Proceedings of the AAAI Conference on Artificial Intelligence},
  volume={40},
  number={30},
  pages={25428--25436},
  year={2026},
  doi={10.1609/aaai.v40i30.39737}
}

@inproceedings{lfrd2,
  title={Learnable Fractional Reaction-Diffusion Dynamics for Under-Display ToF Imaging and Beyond},
  author={Qiao, Xin and Poggi, Matteo and Wei, Xing and Deng, Pengchao and Zhou, Yanhui and Mattoccia, Stefano},
  booktitle={Proceedings of the IEEE/CVF International Conference on Computer Vision},
  pages={6080--6090},
  year={2025}
}

@inproceedings{metzger2023anisotropic,
  title={Guided Depth Super-Resolution by Deep Anisotropic Diffusion},
  author={Metzger, Nando and Daudt, Rodrigo Caye and Schindler, Konrad},
  booktitle={Proceedings of the IEEE/CVF Conference on Computer Vision and Pattern Recognition},
  pages={18237--18246},
  year={2023}
}

@article{rao2023reactiondiffusion,
  title={Encoding Physics to Learn Reaction-Diffusion Processes},
  author={Rao, Chengping and Ren, Pu and Wang, Qi and Buyukozturk, Oral and Sun, Hao and Liu, Yang},
  journal={Nature Machine Intelligence},
  volume={5},
  number={7},
  pages={765--779},
  year={2023},
  doi={10.1038/s42256-023-00685-7}
}

@misc{fractionalattention,
  title={Fractional Neural Attention for Efficient Multiscale Sequence Processing},
  author={Qu, Cheng Kevin and Ly, Andrew and Gong, Pulin},
  year={2025},
  eprint={2511.10208},
  archivePrefix={arXiv},
  primaryClass={cs.LG},
  doi={10.48550/arXiv.2511.10208}
}

@inproceedings{alexnet,
  title={ImageNet Classification with Deep Convolutional Neural Networks},
  author={Krizhevsky, Alex and Sutskever, Ilya and Hinton, Geoffrey E.},
  booktitle={Advances in Neural Information Processing Systems},
  volume={25},
  year={2012}
}

@inproceedings{densenet,
  title={Densely Connected Convolutional Networks},
  author={Huang, Gao and Liu, Zhuang and Van Der Maaten, Laurens and Weinberger, Kilian Q.},
  booktitle={Proceedings of the IEEE Conference on Computer Vision and Pattern Recognition},
  pages={4700--4708},
  year={2017}
}

@misc{mobilenet,
  title={MobileNets: Efficient Convolutional Neural Networks for Mobile Vision Applications},
  author={Howard, Andrew G. and Zhu, Menglong and Chen, Bo and Kalenichenko, Dmitry and Wang, Weijun and Weyand, Tobias and Andreetto, Marco and Adam, Hartwig},
  year={2017},
  eprint={1704.04861},
  archivePrefix={arXiv},
  primaryClass={cs.CV},
  url={https://arxiv.org/abs/1704.04861},
  doi={10.48550/arXiv.1704.04861}
}

@inproceedings{internimage,
  title={InternImage: Exploring Large-Scale Vision Foundation Models with Deformable Convolutions},
  author={Wang, Wenhai and Dai, Jifeng and Chen, Zhe and Huang, Zhenhang and Li, Zhiqi and Zhu, Xizhou and Hu, Xiaowei and Lu, Tong and Lu, Lewei and Li, Hongsheng and Wang, Xiaogang and Qiao, Yu},
  booktitle={Proceedings of the IEEE/CVF Conference on Computer Vision and Pattern Recognition},
  pages={14408--14419},
  year={2023}
}

@inproceedings{xcit,
  title={XCiT: Cross-Covariance Image Transformers},
  author={Ali, Alaaeldin and Touvron, Hugo and Caron, Mathilde and Bojanowski, Piotr and Douze, Matthijs and Joulin, Armand and Laptev, Ivan and Neverova, Natalia and Synnaeve, Gabriel and Verbeek, Jakob and Jegou, Herve},
  booktitle={Advances in Neural Information Processing Systems},
  volume={34},
  pages={20014--20027},
  year={2021}
}

@inproceedings{davit,
  title={DaViT: Dual Attention Vision Transformers},
  author={Ding, Mingyu and Xiao, Bin and Codella, Noel and Luo, Ping and Wang, Jingdong and Yuan, Lu},
  booktitle={European Conference on Computer Vision},
  pages={74--92},
  year={2022},
  doi={10.1007/978-3-031-20053-3_5}
}

@inproceedings{s4nd,
  title={S4ND: Modeling Images and Videos as Multidimensional Signals with State Spaces},
  author={Nguyen, Eric and Goel, Karan and Gu, Albert and Downs, Gordon and Shah, Preey and Dao, Tri and Baccus, Stephen and Re, Christopher},
  booktitle={Advances in Neural Information Processing Systems},
  volume={35},
  pages={2846--2861},
  year={2022},
  doi={10.52202/068431-0206}
}

@misc{qbheat,
  title={Self-Supervised Learning Based on Heat Equation},
  author={Chen, Yinpeng and Dai, Xiyang and Chen, Dongdong and Liu, Mengchen and Yuan, Lu and Liu, Zicheng and Lin, Youzuo},
  year={2022},
  eprint={2211.13228},
  archivePrefix={arXiv},
  primaryClass={cs.CV},
  doi={10.48550/arXiv.2211.13228}
}

@inproceedings{imagenet,
  title={ImageNet: A Large-Scale Hierarchical Image Database},
  author={Deng, Jia and Dong, Wei and Socher, Richard and Li, Li-Jia and Li, Kai and Fei-Fei, Li},
  booktitle={Proceedings of the IEEE Conference on Computer Vision and Pattern Recognition},
  pages={248--255},
  year={2009}
}

@inproceedings{imagenetsketch,
  title={Learning Robust Global Representations by Penalizing Local Predictive Power},
  author={Wang, Haohan and Ge, Songwei and Lipton, Zachary C. and Xing, Eric P.},
  booktitle={Advances in Neural Information Processing Systems},
  volume={32},
  year={2019}
}

@inproceedings{imageneta,
  title={Natural Adversarial Examples},
  author={Hendrycks, Dan and Zhao, Kevin and Basart, Steven and Steinhardt, Jacob and Song, Dawn},
  booktitle={Proceedings of the IEEE/CVF Conference on Computer Vision and Pattern Recognition},
  pages={15262--15271},
  year={2021}
}

@inproceedings{coco,
  title={Microsoft COCO: Common Objects in Context},
  author={Lin, Tsung-Yi and Maire, Michael and Belongie, Serge and Hays, James and Perona, Pietro and Ramanan, Deva and Dollar, Piotr and Zitnick, C. Lawrence},
  booktitle={European Conference on Computer Vision},
  pages={740--755},
  year={2014}
}

@inproceedings{maskrcnn,
  title={Mask R-CNN},
  author={He, Kaiming and Gkioxari, Georgia and Dollar, Piotr and Girshick, Ross},
  booktitle={Proceedings of the IEEE International Conference on Computer Vision},
  pages={2961--2969},
  year={2017},
  doi={10.1109/ICCV.2017.322}
}

@inproceedings{ade20k,
  title={Scene Parsing through ADE20K Dataset},
  author={Zhou, Bolei and Zhao, Hang and Puig, Xavier and Fidler, Sanja and Barriuso, Adela and Torralba, Antonio},
  booktitle={Proceedings of the IEEE Conference on Computer Vision and Pattern Recognition},
  year={2017}
}

@article{metzler2000random,
  title={The Random Walk's Guide to Anomalous Diffusion: A Fractional Dynamics Approach},
  author={Metzler, Ralf and Klafter, Joseph},
  journal={Physics Reports},
  volume={339},
  number={1},
  pages={1--77},
  year={2000},
  doi={10.1016/S0370-1573(00)00070-3}
}

@article{kwasnicki2017ten,
  title={Ten Equivalent Definitions of the Fractional Laplace Operator},
  author={{Kwa{\'s}nicki}, Mateusz},
  journal={Fractional Calculus and Applied Analysis},
  volume={20},
  number={1},
  pages={7--51},
  year={2017},
  doi={10.1515/fca-2017-0002}
}

@book{samorodnitsky1994stable,
  title={Stable Non-Gaussian Random Processes: Stochastic Models with Infinite Variance},
  author={Samorodnitsky, Gennady and Taqqu, Murad S.},
  publisher={Chapman and Hall},
  address={New York},
  year={1994},
  isbn={0412051710}
}

@article{dct,
  title={The Discrete Cosine Transform},
  author={Strang, Gilbert},
  journal={SIAM Review},
  volume={41},
  number={1},
  pages={135--147},
  year={1999},
  doi={10.1137/S0036144598336745}
}

@inproceedings{upernet,
  title={Unified Perceptual Parsing for Scene Understanding},
  author={Xiao, Tete and Liu, Yingcheng and Zhou, Bolei and Jiang, Yuning and Sun, Jian},
  booktitle={Proceedings of the European Conference on Computer Vision},
  pages={418--434},
  year={2018},
  url={https://openaccess.thecvf.com/content_ECCV_2018/html/Tete_Xiao_Unified_Perceptual_Parsing_ECCV_2018_paper.html}
}

@inproceedings{swinir,
  title={SwinIR: Image Restoration Using Swin Transformer},
  author={Liang, Jingyun and Cao, Jiezhang and Sun, Guolei and Zhang, Kai and Van Gool, Luc and Timofte, Radu},
  booktitle={Proceedings of the IEEE/CVF International Conference on Computer Vision Workshops},
  pages={1833--1844},
  year={2021},
  url={https://openaccess.thecvf.com/content/ICCV2021W/AIM/html/Liang_SwinIR_Image_Restoration_Using_Swin_Transformer_ICCVW_2021_paper.html}
}

@article{dncnn,
  title={Beyond a Gaussian Denoiser: Residual Learning of Deep CNN for Image Denoising},
  author={Zhang, Kai and Zuo, Wangmeng and Chen, Yunjin and Meng, Deyu and Zhang, Lei},
  journal={IEEE Transactions on Image Processing},
  volume={26},
  number={7},
  pages={3142--3155},
  year={2017},
  doi={10.1109/TIP.2017.2662206}
}

@article{mcmaster,
  title={Color Demosaicking by Local Directional Interpolation and Nonlocal Adaptive Thresholding},
  author={Zhang, Lei and Wu, Xiaolin and Buades, Antoni and Li, Xin},
  journal={Journal of Electronic Imaging},
  volume={20},
  number={2},
  pages={023016},
  year={2011},
  doi={10.1117/1.3600632}
}

@article{live,
  title={A Statistical Evaluation of Recent Full Reference Image Quality Assessment Algorithms},
  author={Sheikh, Hamid R. and Sabir, Muhammad F. and Bovik, Alan C.},
  journal={IEEE Transactions on Image Processing},
  volume={15},
  number={11},
  pages={3440--3451},
  year={2006},
  doi={10.1109/TIP.2006.881959}
}

\clearpage
\appendix
\numberwithin{figure}{section}
\numberwithin{table}{section}
\section{DCT/IDCT under Neumann Boundary}
\label{app:dct_details}

Consider a bounded rectangular feature domain with spatial indices $0\leq i < H$ and $0\leq j < W$. A homogeneous Neumann boundary condition imposes zero normal derivative at the boundary, so the signal is naturally extended by reflection rather than by periodic wrapping. Cosine bases arise from this even extension and diagonalize the corresponding second-difference operators under Neumann-type boundary choices \citep{dct}. This is the same boundary-motivated spectral choice used by heat-conduction visual propagation \citep{vheat}.

For each channel of a feature map $X\in\mathbb{R}^{C\times H\times W}$, the two-dimensional transform is implemented separably. Let $D_H$ and $D_W$ denote orthonormal one-dimensional DCT matrices along height and width. The forward and inverse transforms can be written as
\begin{equation}
\widehat{X}_c = D_H X_c D_W^\top,
\qquad
X_c = D_H^\top \widehat{X}_c D_W,
\label{eq:appendix_dct_pair}
\end{equation}
for channel $c$. The discrete frequency grid is indexed by $(m,n)$, and TPO uses the squared frequency magnitude $\rho_{mn}=\omega_m^2+\omega_n^2$. Substituting this grid into the stable transfer response gives the discrete Gaussian and Cauchy factors in \eqref{eq:discrete_responses}, which are then applied elementwise to DCT coefficients.

\section{Separable Matrix-DCT and Fused TPO}
\label{app:matrix_dct}

The current implementation uses an explicit matrix-DCT backend. For each channel, the horizontal and vertical transforms are performed by multiplying with $D_W^\top$ and $D_H$, followed by the corresponding inverse multiplications. This separable implementation costs $O(C(H^2W+HW^2))$ for a feature map with $C$ channels. For square feature maps with $N=HW$, this becomes $O(CN^{1.5})$, and the spatial scaling with respect to $N$ is $O(N^{1.5})$ when the channel width is held fixed.

TPO does not compute separate inverse transforms for the Gaussian and Cauchy branches. Since the mixing coefficient $\lambda(X)$ is channel-wise and spatially shared, no spatial routing tensor is constructed, and the two transfer responses can be fused before the inverse DCT:
\begin{equation}
\operatorname{IDCT}_{2D}\!\left(
H_{\mathrm{TPO}}(X,\rho)
\odot
\operatorname{DCT}_{2D}(X)
\right),
\quad
H_{\mathrm{TPO}}(X,\rho)
=
\lambda(X)H_G(\rho)+(1-\lambda(X))H_C(\rho).
\label{eq:appendix_fused_tailprop}
\end{equation}
This is the fused implementation of TPO and is algebraically equivalent to applying the two spectral branches separately and mixing their real-valued outputs. The local workspace includes fused-versus-explicit equivalence tests, and the recorded FP32 maximum output discrepancy is on the order of $10^{-7}$ in the corresponding analysis note.

\section{TailProp Architecture Configurations}
\label{app:tailprop_arch_configs}

TailProp mirrors the four-stage vHeat-style macro architecture: a patch-size-4 stem, stage resolutions of $56^2$, $28^2$, $14^2$, and $7^2$ for 224-pixel inputs, and a final LayerNorm2d + adaptive average pooling + linear classifier head. The scale-specific changes are the stage depths, channel widths, drop-path rate, and the post-norm / layer-scale choice.

\begin{center}
  \captionsetup{type=table}
  \caption{TailProp scale configurations used in the paper. The classifier head is dataset-specific and follows the ImageNet-1K class convention for main classification experiments.}
  \label{tab:tailprop_arch_configs}
  \scriptsize
  \resizebox{\textwidth}{!}{%
  \begin{tabular}{lcccccc}
    \toprule
    Model & Depths & Dims & Stage res. & Drop path & Post-norm & Layer scale \\
    \midrule
    TailProp-T & [2, 2, 6, 2] & [96, 192, 384, 768] & [56$^2$, 28$^2$, 14$^2$, 7$^2$] & 0.1 & No & -- \\
    TailProp-S & [2, 2, 18, 2] & [96, 192, 384, 768] & [56$^2$, 28$^2$, 14$^2$, 7$^2$] & 0.3 & Yes & 1e-5 \\
    TailProp-B & [2, 2, 18, 2] & [128, 256, 512, 1024] & [56$^2$, 28$^2$, 14$^2$, 7$^2$] & 0.5 & Yes & 1e-5 \\
    \bottomrule
  \end{tabular}}
\end{center}

\section{Additional Experimental Details}
\label{app:experimental_details}

\subsection{Image Classification}
The ImageNet-1K classification experiments use the Hugging Face parquet snapshot of ILSVRC-2012 with 294 training shards, 14 validation shards, 1,281,167 training images, 50,000 validation images, and 1,000 classes. The training entry point is \texttt{scripts/train\_imagenet1k\_tailprop\_ddp.py}; launch manifests are written by the TailProp ImageNet-1K launcher scripts before each run. Unless otherwise stated, TailProp-T/S/B use 224-pixel inputs, BF16 autocast, AdamW with weight decay 0.08, cosine learning-rate decay, 20 warmup epochs, base learning rate $5\times10^{-4}$ scaled by effective global batch size over 512, minimum learning rate $5\times10^{-6}$, gradient clipping at 5.0, label smoothing 0.1, mixup 0.8, cutmix 1.0, color jitter 0.4, RandAugment \texttt{rand-m9-mstd0.5-inc1}, and random erasing 0.25. Evaluation uses resize-256 and center-crop-224 preprocessing. Experiment configurations and throughput/FPS measurements are reported for NVIDIA A100 80GB GPUs, matching the hardware convention of vHeat and WaveFormer. The main 300-epoch runs keep \texttt{last.pt} and \texttt{best.pt}; periodic checkpoints are used only when enabled by the launch manifest.

\subsection{Object Detection and Instance Segmentation}
For MS COCO 2017, we initialize Mask R-CNN from the corresponding ImageNet-1K classification-pretrained backbone and keep the detector head, image scale policy, optimizer, and augmentation matched across backbones. TailProp-T uses depths $(2,2,6,2)$, widths $96$, drop-path $0.1$, and no post-normalization; TailProp-S uses depths $(2,2,18,2)$, widths $96$, drop-path $0.3$, post-normalization, and layer scale $10^{-5}$; TailProp-B uses depths $(2,2,18,2)$, widths $128$, drop-path $0.5$, post-normalization, and layer scale $10^{-5}$. We follow the standard Mask R-CNN + FPN recipe with AdamW ($\text{lr}=10^{-4}$, betas $(0.9,0.999)$, weight decay $0.05$), a 1000-iteration linear warmup, and MultiStepLR milestones at $[8,11]$ for the 1x schedule and $[27,33]$ for the 3x schedule. The 1x schedule runs for 12 epochs and the 3x schedule for 36 epochs; the 3x augmentation uses DETR/Sparse R-CNN style RandomChoiceResize over $480$--$800$ with the intermediate crop branch \texttt{RandomCrop(384,600)}. We report box AP, mask AP, FPS, and FLOPs for both schedules. Following the convention of vHeat and WaveFormer, FLOPs are computed at input size $1280\times800$, and FPS is measured on NVIDIA A100 80GB GPUs.

\subsection{Semantic Segmentation}
For ADE20K semantic segmentation, we initialize UPerNet from the corresponding ImageNet-1K classification-pretrained backbone and keep the decoder, crop size, optimizer, schedule, multi-scale testing choice, and augmentation matched across compared methods. TailProp-T uses depths $(2,2,6,2)$, widths $96$, drop-path $0.3$, and no post-normalization; TailProp-S uses depths $(2,2,18,2)$, widths $96$, drop-path $0.4$, post-normalization, and layer scale $10^{-5}$; TailProp-B uses depths $(2,2,18,2)$, widths $128$, drop-path $0.5$, post-normalization, and layer scale $10^{-5}$. We follow the standard 160k UPerNet recipe with crop size $512\times512$, AdamW ($\text{lr}=6\times10^{-5}$, betas $(0.9,0.999)$, weight decay $0.01$), a 1500-iteration linear warmup, PolyLR with power $1.0$, train batch size 2, validation batch size 1, and slide testing with stride $(341,341)$. The main metric is mIoU. Following the vHeat and WaveFormer convention, FLOPs are computed at input size $512\times512$, and FPS is measured on NVIDIA A100 80GB GPUs.

\subsection{Robustness Evaluation}
For ImageNet-Sketch and ImageNet-A, we evaluate the ImageNet-1K-pretrained classifiers without additional finetuning and report Top-1 accuracy. ImageNet-Sketch is evaluated over the complete 1,000 ImageNet classes, whereas ImageNet-A is evaluated on its official 200-class naturally adversarial subset using the corresponding ImageNet class indices. The evaluation uses the same normalization and classifier-head convention as the ImageNet-1K validation pipeline.

\subsection{Cross-Backbone Generalization}
For cross-backbone generalization, we follow the official SwinIR restoration protocol and instantiate TailPropIR by replacing the token-mixing modules with TPOs while keeping the reconstruction head and task losses matched. The training corpus is DFWB, the standard SwinIR denoising/JPEG corpus built from DIV2K, Flickr2K, BSD500, and WED. The denoising configuration uses a SwinIRNet backbone with $\text{img\_size}=128$, window size 8, six residual groups of six blocks, embed dimension 180, paired random crop size 128, Adam ($\text{lr}=2\times10^{-4}$, betas $(0.9,0.999)$), batch size 1, and 1.6M iterations with multi-step decays at 800k, 1.2M, 1.4M, 1.5M, and 1.6M. The JPEG artifact reduction configuration uses the same optimizer and iteration budget with SwinIRNet $\text{img\_size}=126$, window size 7, paired random crop size 126, and JPEG quality factor $q=40$. We evaluate grayscale denoising on Set12 with $\sigma=15$, color denoising on McMaster with $\sigma=15$, and JPEG artifact reduction on LIVE1 with $q=40$, reporting PSNR under the corresponding SwinIR-style test pipelines.

\section{Additional Ablations}
\label{app:ablation_details}

\subsection{Control Definitions}
All controls keep the hierarchical backbone, classifier head, image resolution, optimizer, augmentation, and training schedule matched unless explicitly stated, so the comparison isolates the propagation basis and adaptive mixture used by TPO. Let $G(X)$ and $C(X)$ denote the Gaussian and Cauchy propagation responses of a feature map $X$. Gaussian-only sets $Y=G(X)$, and Cauchy-only sets $Y=C(X)$. Fixed G+C keeps both bases but uses a fixed coefficient,
\[
Y=0.5\,G(X)+0.5\,C(X).
\]
Learnable G+C replaces the fixed coefficient with an input-independent learnable gate,
\[
\boldsymbol{\lambda}=\sigma(\boldsymbol{a}),\qquad
Y=\boldsymbol{\lambda}\odot G(X)+(1-\boldsymbol{\lambda})\odot C(X),
\]
where $\boldsymbol{a}$ is a learned channel-wise parameter in each TPO layer. It is independent of $X$ and is shared by all samples and spatial positions. Its final learned weights average to Gaussian = 0.53 and Cauchy = 0.47. TailProp instead predicts the mixing weights from the input,
\[
\boldsymbol{\lambda}(X)=\sigma(g(\mathrm{GAP}(X))),\qquad
Y=\boldsymbol{\lambda}(X)\odot G(X)+[1-\boldsymbol{\lambda}(X)]\odot C(X),
\]
where $\boldsymbol{\lambda}(X)$ is channel-wise and spatially shared for each sample in each TPO layer. Thus Learnable G+C learns a single dataset-level mixing preference, whereas TailProp predicts input-conditioned mixing weights. Dual Gaussian keeps the dual-branch and input-adaptive routing structure but replaces the Cauchy basis with a second Gaussian branch. Adaptive $\alpha$ replaces the explicit Gaussian--Cauchy pair with one content-adaptive fractional stable propagation order.

\section{Implementation Details}
\label{app:implementation_details}

The implementation follows the operator definition in \Secref{sec:tailprop}. Each TPO Block first applies a depth-wise $3\times3$ convolution and a linear projection, after which the projected feature is split into propagation and gating branches. The propagation branch applies TPO and LayerNorm, whereas the gating branch is activated by SiLU; the two branches are then combined by element-wise modulation before the final output projection. The block is wrapped by a vHeat-style residual layer with stochastic depth, an MLP branch, optional post-normalization, and optional layer-scale parameters, depending on the TailProp-T/S/B configuration.

The TPO module uses the fused DCT path described in \Appref{app:matrix_dct}. In the input-adaptive variant, the Gaussian and Cauchy responses are formed on the DCT Laplacian grid, mixed into one channel-wise response tensor, multiplied with the DCT coefficients, and then passed through a single inverse DCT. The content-conditioned gate is implemented as GAP--MLP--Sigmoid with a reduction ratio of 8, producing a per-sample, per-channel, spatially shared $\boldsymbol{\lambda}(X)$. The Gaussian and Cauchy propagation scales are learnable positive scalars parameterized by a softplus transform with a small positive offset. The DCT grid is cached by feature resolution and device, and the fused spectral multiplication can compute the response in FP32 before casting back to the transform dtype, which stabilizes BF16 mixed-precision training.

Training and evaluation scripts use PyTorch DistributedDataParallel with NCCL, DistributedSampler for train and validation splits, BF16 autocast by default, rank-reduced metrics, finite-loss checks, optional finite-gradient checks, and JSONL heartbeat logging. Checkpoints store the model, optimizer, scheduler, AMP scaler when enabled, epoch/global step, best metrics, training history, and RNG states for resume. The launch scripts also write command, metadata, configuration hash, dataset split hash, model configuration, optimizer/scheduler settings, throughput metadata, and checkpoint policy into each run directory. Local verification used Python 3.10.18, PyTorch 2.8.0+cu128, torchvision 0.23.0+cu128, and timm 1.0.28.

\section{Additional Mechanism Diagnostics}
\label{app:mechanism_diagnostics}

For each hooked TPO block, the gate tensor $\boldsymbol{\lambda}(X)$ is channel-wise and spatially shared. We first average it over channels and spatial positions to obtain a per-sample stage value, then pool those values across all hooked blocks in the same stage and across the audited validation samples. The effective Cauchy contribution is computed from the learned transfer responses as
\[
\frac{\sum ((1-\boldsymbol{\lambda})\odot h_C \odot E)}{\sum (h_{\mathrm{mix}}\odot E)},
\qquad
h_{\mathrm{mix}}=\boldsymbol{\lambda}\odot h_G + (1-\boldsymbol{\lambda})\odot h_C,
\]
where $E = \lvert \operatorname{DCT}(X)\rvert^2$. The companion response-only Cauchy score is the mean of $(1-\boldsymbol{\lambda})\odot h_C / h_{\mathrm{mix}}$ over channels and spatial positions before the energy weighting above. HF energy is the normalized DCT energy above the radius threshold $\rho \ge 0.5$, namely $\sum_{\rho\ge 0.5} E / \sum E$. Figure G.1 reports the seed-mean of these stage-wise summaries over seeds 2027, 42, and 3407.

\begin{center}
  \centering
  \captionsetup{type=figure}
  \includegraphics[width=0.72\textwidth]{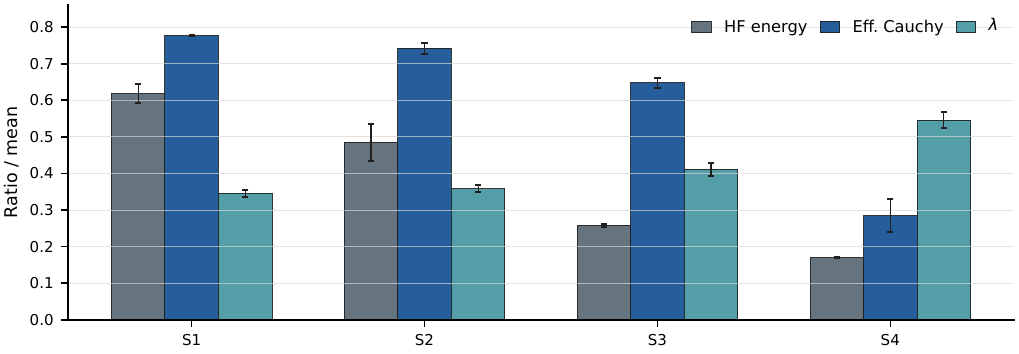}
  \caption{Stage-wise TPO mechanism diagnostics across S1--S4, summarizing gate preference, effective contribution, and feature-frequency behavior.}
  \label{fig:appendix_stagewise_mechanism}
\end{center}

\section{Additional ERF Visualizations}
\label{app:erf_visualizations}

\begin{center}
  \centering
  \captionsetup{type=figure}
  \includegraphics[width=0.82\textwidth]{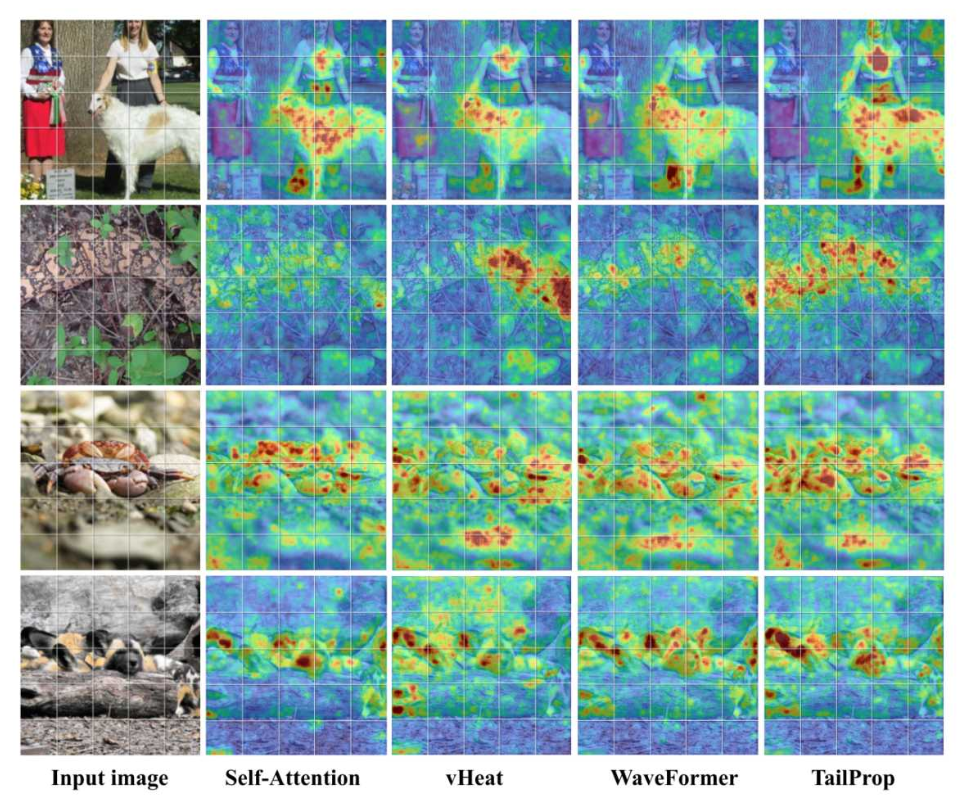}
  \caption{Representative additional ERF comparisons for four ImageNet validation cases. Each group follows the same four-model ERF protocol as Fig.~1, comparing Self-Attention, vHeat, WaveFormer, and TailProp.}
  \label{fig:appendix_erf_h1}
\end{center}

\section{Additional Controlled TPO Visualizations}
\label{app:controlled_tpo_visualizations}
This appendix complements \Figref{fig:gate_cases} and \Figref{fig:controlled_tpo} with a deterministic multi-source view of TailProp propagation using the same trained TailProp checkpoint with learned Stage-3 propagation scales.

\subsection{Multi-source Controlled TPO Propagation}
\label{app:multisource_tpo}
\begin{figure}[htbp]
  \centering
  \includegraphics[width=\textwidth]{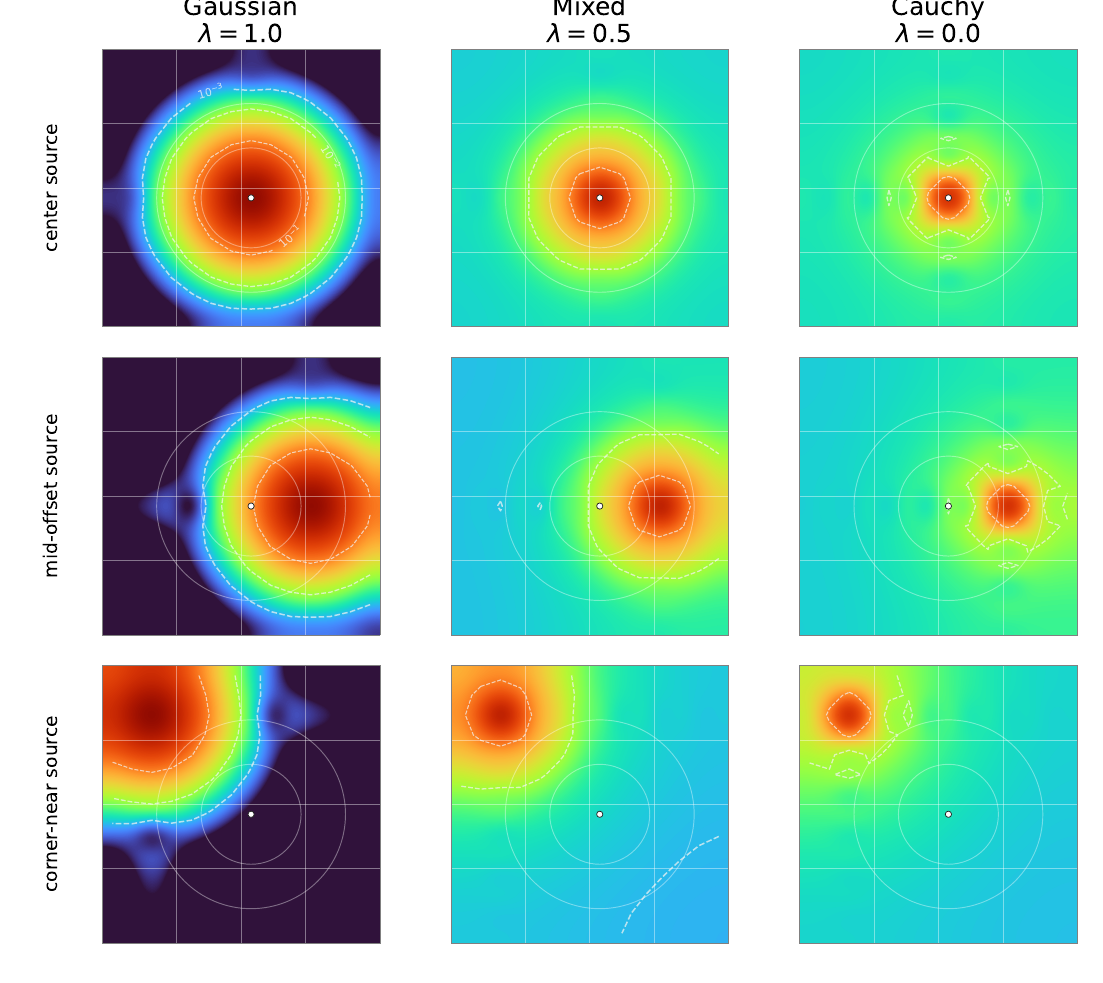}
  \caption{Controlled TPO responses for center, mid-offset, and corner-near source locations using the same trained TailProp checkpoint with learned Stage-3 propagation scales. Each row fixes the source and varies only the Gaussian--Cauchy mixing coefficient; all panels use one peak-normalized log scale with shared contour thresholds.}
  \label{fig:appendix_tpo_multisource}
\end{figure}

\Figref{fig:appendix_tpo_multisource} extends the controlled impulse response in \Figref{fig:controlled_tpo} to three deterministic source locations while keeping the learned Stage-3 propagation scales fixed.

\end{document}